\documentclass[conference]{IEEEtran}
\IEEEoverridecommandlockouts
\usepackage{cite}
\usepackage{amsmath,amssymb,amsfonts}
\usepackage{algorithmic}
\usepackage{graphicx}
\usepackage{textcomp}
\usepackage{xcolor}
\def\BibTeX{{\rm B\kern-.05em{\sc i\kern-.025em b}\kern-.08em
    T\kern-.1667em\lower.7ex\hbox{E}\kern-.125emX}}

\newtheorem{definition}{Definition}

\usepackage{bm}
\usepackage{amsmath}

\begin{document}

\title{BiteNet: Bidirectional Temporal Encoder Network to Predict Medical Outcomes}
\author{
\IEEEauthorblockN{
    Xueping Peng\IEEEauthorrefmark{1},
    Guodong Long\IEEEauthorrefmark{1},
    Tao Shen\IEEEauthorrefmark{1},
    Sen Wang\IEEEauthorrefmark{2},
    Jing Jiang\IEEEauthorrefmark{1},
    Chengqi Zhang\IEEEauthorrefmark{1}
}
\IEEEauthorblockA{
    \IEEEauthorrefmark{1} Australian Artificial Intelligence Institute, FEIT, University of Technology Sydney, Australia \\ 
    \IEEEauthorrefmark{2} School of Information Technology and Electrical Engineering, The University of Queensland, Australia \\
    Email: \{xueping.peng, guodong.long\}@uts.edu.au, Tao.Shen@student.uts.edu.au, \\ sen.wang@uq.edu.au, \{jing.jiang, chengqi.zhang\}@uts.edu.au}
}


\maketitle

\begin{abstract}
Electronic health records (EHRs) are longitudinal records of a patient's interactions with healthcare systems. A patient's EHR data is organized as a three-level hierarchy from top to bottom: patient journey - all the experiences of diagnoses and treatments over a period of time; individual visit - a set of medical codes in a particular visit; and medical code - a specific record in the form of medical codes. As EHRs begin to amass in millions, the potential benefits, which these data might hold for medical research and medical outcome prediction, are staggering - including, for example, predicting future admissions to hospitals, diagnosing illnesses or determining the efficacy of medical treatments. Each of these analytics tasks requires a domain knowledge extraction method to transform the hierarchical patient journey into a vector representation for further prediction procedure. The representations should embed a sequence of visits and a set of medical codes with a specific timestamp, which are crucial to any downstream prediction tasks. Hence, expressively powerful representations are appealing to boost learning performance. To this end, we propose a novel self-attention mechanism that captures the contextual dependency and temporal relationships within a patient's healthcare journey. An end-to-end bidirectional temporal encoder network (BiteNet) then learns representations of the patient's journeys, based solely on the proposed attention mechanism. We have evaluated the effectiveness of our methods on two supervised prediction and two unsupervised clustering tasks with a real-world EHR dataset. The empirical results demonstrate the proposed BiteNet model produces higher-quality representations than state-of-the-art baseline methods.
\end{abstract}


\section{Introduction}
\begin{figure}[t]
	\setlength{\belowcaptionskip}{-15pt}
	\setlength{\abovecaptionskip}{5pt} 
	\centering
	\scalebox{0.45}{\includegraphics{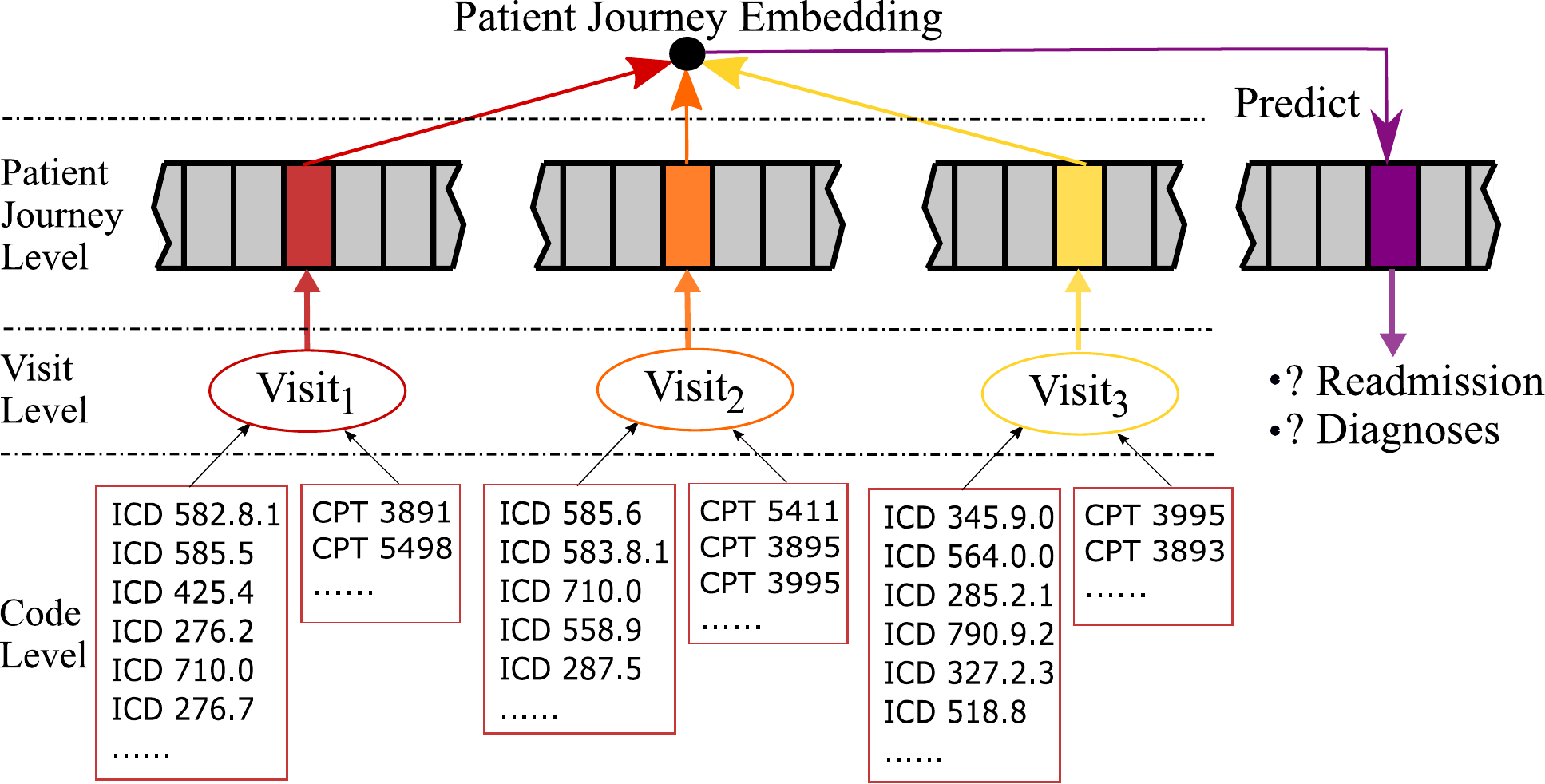}}
	\caption{\small An example of a patient's journey that is hierarchically structured in three levels: patient journey, individual visit and medical code.}
	\label{ehr}
\end{figure}

Healthcare information systems store huge volumes of electronic health records (EHRs) that contain detailed visit information about patients over a period of time~\cite{Shickel_2018}. The data is structured in three levels from top to bottom: the patient journey, the individual visit and the medical code. Fig.~\ref{ehr} provides a typical example of this structure. An anonymous patient visits his/her doctor, a pathology lab and is admitted to the hospital on different days. The procedures and diagnoses performed at each of these visits are recorded as industry-standard medical codes. Each medical code, i.e. International Classification of Diseases (ICD) and Current Procedure Terminology (CPT), at the lowest level, records an independent observation while the set of codes at a higher level can depict the medical conditions of a patient at a given time point. At the top level, all occurrences of medical events at different time-stamps are chained together as a patient journey, which offers more informative details. Predicting sequential medical outcomes based on a patient's journey, such as hospital re-admissions and diagnoses, is a core research task that significantly benefits for healthcare management by hospitals and governments. For example, re-admission statistics could be used to measure the quality of care; Diagnoses can be used to understand more fully a patient's problems and relevant medical research~\cite{Rajkomar_Google_2018}. However, researchers have encountered many challenges in their attempts to represent patient journeys and predict medical outcomes from EHR data with the characteristics of temporality, high-dimensionality and irregularity~\cite{Qiao_2018}.

Recurrent neural networks (RNNs) have been widely used to analyze sequential data, unsurprisingly including medical events modelling for clinical prediction~\cite{Choi_2016,Choi_Bahadori_2017,choi2018mime,ma2017dipole,Qiao_2018}. For example, Choi et al.~\cite{Choi_2016,choi2018mime} proposed a multi-level representation learning, which integrates visits and medical concepts based on visit sequences and the co-occurrence of medical concepts. They indirectly exploited an RNN to embed the visit sequences into a patient representation for downstream prediction tasks. Some other research works directly employed RNNs to model time-ordered patient visits for predicting diagnoses~\cite{choi2016retain,Choi_Bahadori_2017,ma2017dipole,Qiao_2018}. 
However, when the length of the patient visit sequence grows, such RNN-based models are restricted by the less expressive power of RNNs, such as vanishing gradient and forgetfulness. 
However, such RNN-based models are constrained by forgetfulness, i.e., their predictive power drops significantly when the sequence of patient visits grows too long. 
To memorize historical records, LSTM~\cite{hochreiter1997long} and GRU~\cite{cho2014learning} have been developed to utilize memory and gate mechanism for mitigating these issues.  To go further, Song et al.~\cite{song2018attend} proposed to utilise attention mechanism in a deep framework to model sequential medical events. 
It is worth noting that sequences of medical events are often found to be lengthy~\cite{Johnson_2016}, especially when a patient suffers from chronic disease. Hence, due to the restricted ability of RNNs for long-term dependency modeling \cite{vaswani2017attention}, the traditional RNNs, even with memory cells and gates, usually underperform in the cases of a long sequence of medical events. In light of this, a neural model that can overcome the performance bottleneck of RNN-based models is particularly desirable for medical predictions based on longitudinal EHR data. 



Most recently, attention mechanisms~\cite{Bahdanau_2014} have sprung to the fore as effective integrations with RNNs for modeling sequential EHR data~\cite{choi2016retain,ma2017dipole,Rajkomar_Google_2018,niu2020multichannel}. So far, these approaches have shown satisfactory prediction accuracy, but some argue that the power of attention in an RNN is limited by weaknesses in the RNN itself \cite{vaswani2017attention,shen2018disan}. In particular, Vaswani et al.~\cite{vaswani2017attention} used a sole attention mechanism, i.e., multi-head attention and self-attention, to construct a sequence-to-sequence model for neural machine translation tasks and achieved a state-of-the-art quality score. And according to  Shen et al.~\cite{shen2018disan}, self-attention mechanism allows for more flexibility in sequence lengths than RNNs and is more task/data-driven when modeling contextual dependencies. Unlike recurrent models, attention procedure is easy to compute and the computation can also be significantly accelerated with distributed/parallel computing schemes. 
For example, Song et al.~\cite{song2018attend} proposed to employ 1D CNN \cite{Kim2014-pk} to model local context and use attention mechanism \cite{vaswani2017attention} to capture long-term dependency for sequential medical events. 
However, when applied to EHR data instead of regular sequential data (e.g., natural language text), the current attention models cannot appropriately deal with some aspects of EHR data, such as arbitrary time-stamps and hierarchical data format. 
Hence, to the best of our knowledge, a neural network-based entirely on attention has never been designed for analytics with EHR data. 

To bridge the gap in this literature and address some of the open issues listed above, we propose a novel attention mechanism called \textbf{Mas}ked \textbf{Enc}oder (MasEnc) for temporal context fusion. It uses self-attention to capture contextual information and temporal dependencies between a patient's visits. 
Then, we propose an end-to-end neural network, called \textbf{Bi}directional \textbf{t}emporal \textbf{e}ncoder \textbf{Net}work (BiteNet), to predict medical outcomes by leveraging a learned representation of the patient journey,  where the representation is generated solely by the proposed attention mechanism, MasEnc. BiteNet constructs a multi-level self-attention network to represent visits and patient journeys simultaneously, using attention pooling and stacked MasEnc layers. It is worth noting that, compared to the existed RNN-based methods, BiteNet can yield better prediction performance for long sequences of medical records. 

Experiments conducted on two supervised prediction and two unsupervised clustering tasks with real-world EHR datasets demonstrate that the proposed BiteNet model is superior to prior state-of-the-art baseline methods. 

To summarize, our main contributions are:
\begin{itemize}
	\item a novel attention mechanism, called MasEnc, that uses self-attention to capture the contextual information and long-term dependencies between patients' visits;
	\item an end-to-end neural network, called BiteNet, that predicts medical outcomes using a learned representation of a patient journey based solely on the proposed attention mechanism;
	\item evaluations of the proposed model on a real-world dataset with supervised and unsupervised tasks, demonstrating that the BiteNet is superior to all the comparative methods.
\end{itemize}

\section{Related Work}\label{Relat}
Applying deep learning to healthcare analytical tasks has recently attracted enormous interest in healthcare informatics. This section reviews two types of related studies, which are Medical Concept Embedding and patient journey embedding in EHRs. 
\subsection{Medical Concept Embedding}
Word representation models are first introduced from ~\cite{jha2018interpretable,Mikolov_2013,Mikolov_2013_b}. Borrowing ideas from those models, researchers in the healthcare domain have recently explored the possibility of creating representations of medical concepts~\cite{peng2019attentive}. Much of this research has focused on the Skip-gram model. For example, Minarro-Gimnez et al.~\cite{Minarro_2014} directly applied Skip-gram to learn representations of medical text, and Vine et al.~\cite{De_Vine_2014} did the same for UMLS medical concepts. Choi et al.~\cite{Choi_AMIA_2016} went a step further and used the Skip-gram model to learn medical concept embeddings from different data sources, including medical journals, medical claims, and clinical narratives. In other work~\cite{Choi_2016}, Choi et al. developed the Med2Vec model based on Skip-gram to learn concept-level and visit-level representations simultaneously. The shortcoming of all these models is that they view EHRs as documents in the NLP sense, which means that temporal information is ignored. 

\subsection{Patient Journey Embedding}
Applying deep learning to healthcare analytical tasks has recently attracted enormous interest in healthcare informatics. RNN has been widely used to model medical events in sequential EHR data for clinical prediction tasks~\cite{Choi_2016,Choi_Bahadori_2017,choi2018mime,ma2017dipole,Qiao_2018,Ma2018-gu,Peng2019TemporalSN,peng2020self}. Choi et al.~ \cite{Choi_2016,choi2018mime} indirectly exploit an RNN to embed the visit sequences into a patient representation by multi-level representation learning to integrate visits and medical codes based on visit sequences and the co-occurrence of medical codes. Other research works have, however, used RNNs directly to model time-ordered patient visits for predicting diagnoses~\cite{choi2016retain,Choi_Bahadori_2017,ma2017dipole,Qiao_2018,Ma2018-gu,Ma2018-ao,ma2018health}. CNN has been exploited to represent a patient journey  in other way. For example, Nguyen et al.~\cite{nguyen2016deepr} transform a record into a sequence of discrete elements separated by coded time gaps, and then employ CNN to detect and combine predictive local clinical motifs to stratify the risk. Song et al.~\cite{song2018attend} use CNN in code level to learn visit embedding. These RNN- and CNN-based models follow ideas of processing sentences~\cite{Kim2014-pk} in documents from NLP to treat a patient journey as a document and a visit as a sentence, which only has a sequential relationship, while two arbitrary visits in one patient journey may be separated by different time intervals, an important factor in longitudinal studies. Attention-based neural networks have been exploited successfully in healthcare tasks to model sequential EHR data~\cite{choi2016retain,ma2017dipole,Rajkomar_Google_2018,song2018attend} and have been shown to improve predictive performance.

\section{Background}\label{Background}

This section starts with definitions of several important concepts and tasks in the paper. The remainder mainly focuses on how we adapt embeddings and attention mechanism from natural language processing (NLP) to EHR data and analytics with patient journeys. 

\subsection{Definitions and Notations}

\begin{definition}[Medical Code] 
A medical code is a term or entry to describe a diagnosis, procedure, medication, or laboratory test administered to a patient. 
A set of medical codes is formally denoted as  $X=\{x_1, x_2,\dots, x_{|X|}\}$, where $|X|$ is the total number of unique medical codes in the EHR data.
\end{definition}

\begin{definition}[Visit]\label{visit}
A visit is a hospital stay from admission to discharge with an admission time stamp. A visit is denoted as $V_{i} = <x_i, t_i>$, where  $x_i = [x_{i1},x_{i2},...,x_{ik}]$, $i$ is the $i$-th visit, $t_i$ is the admission time of the $i$-th visit,  $k$ is the total number of medical codes in a visit.
\end{definition}

\begin{definition}[Patient Journey] 
A patient journey consists of a sequence of visits over time, denoted as $J = [V_{1},V_{2},...,V_{n}]$, where $n$ is the total number of visits for the patient. 
\end{definition}

\begin{definition}[Temporal Interval] \label{interval}
Temporal interval refers to a difference in days between admission time $t_i$ of the $i$-th visit and admission time $t_1$ of the first visit in a patient journey, denoted as $p_{i} = |t_{i} - t_1|$, where $i = 1, \dots, n$.
\end{definition}

\begin{definition}[Task] 
Given a set of patient journeys \{$J_l\}_{l=1, \dots}$, the task is to predict medical outcomes.
\end{definition}

We choose to predict diagnoses and future hospital re-admissions as examples. However, many other outcomes could be predicted.

In addition, a patient's medical data means a stored and chronological sequence of {$n$} visits in a patient journey $J_l$. To reduce clutter, we omit the subscript ($l$) indicating $l$-th patient, when discussing a single patient journey.

Table~\ref{tab_notes} summarizes notations we will use throughout the paper.

\renewcommand{\arraystretch}{1.4}
\begin{table}[htbp]\small
\caption{\small Notations for BiteNet.}\label{tab_notes}
\centering
	\scalebox{1}{
        \begin{tabular}{|c|l|}
        \hline
        \textbf{Notation} & \textbf{Description} \\
        \hline
        $X$      & Set of unique medical codes  \\ \hline
        $|X|$     & The size of unique medical concept \\ \hline
        $V_{i}$        & The \textit{i}-th visit of the patient\\ \hline
        $v_{i}$        & The representation of \textit{t}-th visit of the patient\\ \hline 
        $\bm{v}$   & Sequence of $n$ visit embeddings of the patient\\ \hline
        ${x}_{i}$     & Set of medical codes in $V_{i}$  \\ \hline
       $\bm{x}$ &  Sequence of medical codes, $[x_1, x_2, ..., x_n]$\\ \hline
        ${e}_{i}$ & Set of medical code embeddings in $\bm{x}_{i}$  \\ \hline 
        $\bm{e}$ &   \shortstack[l]{A sequence of medical code embeddings, \\ $[e_1, e_2, ..., e_n]$}\\ \hline
        ${J}$       & \shortstack[l]{A  patient  journey  consisting of  a  sequence  of \\ visits  over  time}\\ \hline
        $d$       & The dimension of medical code embedding\\ 
        \hline
\end{tabular}}
\end{table}

\subsection{Medical Code Embedding}\label{code_embedding}

The concept of word embedding was introduced to medical analytics by Mikolov et al.~\cite{Mikolov_2013_b,Mikolov_2013} as a way to learn low-dimensional real-valued distributed vector representations of medical codes for downstream tasks instead of using discrete medical codes. This makes medical code embedding a fundamental processing unit for learning EHRs in a deep neural network. Formally, given a sequence or set of medical concepts $\textit{\textbf{x}}=[x_1, x_2, ..., x_n]\in \mathbb{R}^{|X| \times n}$, where $x_i$ is a one-hot vector, and $n$ is the sequence length.
In the NLP literature, a word embedding method like word2vec \cite{Mikolov_2013_b,Mikolov_2013} converts a sequence of one-hot vectors into a sequence of low-dimensional vectors $\textit{\textbf{e}} = [e_1, e_2, ..., e_n] \in \mathbb{R}^{d \times n}$, where $d$ is the embedding dimension of $e_i$. This process can be formally written as $\textbf{\textit{e}} = W^{(e)}\textit{\textbf{x}}$, where $W^{(e)} \in \mathbb{R}^{d \times |X|}$ is the embedding weight matrix, which can be fine-tuned during the training phase. 

\subsection{Visit Embedding}

As stated in Definition~\ref{visit}, a visit consists of a set of medical codes. 
Therefore, a visit can be represented by the set of medical codes embedded with real-valued dense vectors. A straightforward approach to learning this representation $v_i$ is to sum the embeddings of medical codes in the visit, which is written as
\begin{equation}
\label{visit_embedding}
v_i = \sum_{e_{ik} \in {e}_{i}} e_{ik},
\end{equation}
where ${e}_{i}$ is the set of medical code embeddings in the $i$-th visit, and $e_{ik}$ is the $k$-th code embedding in ${e}_{i}$. A visit can also be represented as a real-valued dense vector with a more advanced method, such as self-attention and attention pooling discussed below.

\subsection{Attention Mechanism}
\subsubsection{Vanilla Attention}

Given a patient journey consisting of a sequence of visits  $\bm{v} = [v_1, v_2, ...,v_m]$ ($\bm{v} \in \mathbb{R}^{m \times d}$) and a vector representation of a query $q \in \mathbb{R}^d$, vanilla attention~\cite{Bahdanau_2014} computes the alignment score between $q$ and each visit $v_i$ using a compatibility function $f(v_i, q)$. A softmax function then transforms the alignment scores $\alpha \in \mathbb{R}^n$ to a probability distribution $p(z|\bm{v}, q)$, where \textit{z} is an indicator of which visit is important to \textit{q}. A large $p(z = i|\bm{v}, q)$ means that $v_i$ contributes important information to $q$. This attention process can be formalized as
\begin{equation}
\alpha = [f(v_i, q)]_{i=1}^n,
\end{equation}
\begin{equation}
p(z|\bm{v}, q) = softmax(\alpha).
\end{equation}
The output $\bm{s}$ denotes the weighted visits according to their importance, i.e.,

\begin{equation}
\bm{s} = p(z|\bm{v}, q)\odot \bm{v}.
\end{equation}

Additive attention is a commonly-used attention mechanism~\cite{Bahdanau_2014,shang2015neural}, where a compatibility function $f(\cdot)$ is parameterized by a multi-layer perceptron (MLP), i.e., 
\begin{equation}
f(v_i, q) = w^T \sigma (W^{(1)}v_i+W^{(2)}q + b^{(1)})+b,
\end{equation}
where $W^{(1)} \in \mathbb{R}^{d\times d}$, $W^{(2)} \in \mathbb{R}^{d\times d}, w \in \mathbb{R}^d$ are learnable parameters, and $\sigma(\cdot)$ is an activation function. 

Additive attention tends to be memory-intensive with long runtimes. However, it usually produces better representations for downstream tasks than multiplicative attention~\cite{sukhbaatar2015end,rush2015neural}, with cosine similarity as the compatibility function. 

\subsubsection{Attention Pooling}\label{attn_pool}
Attention Pooling~\cite{lin2017structured,liu2016learning} explores the importance of each individual visit within an entire patient journey, given a specific task. It works by compressing a sequence of visit embeddings from a patient journey into a single context-aware vector representation for downstream classification or regression. Formally, {$q$} is removed from the common compatibility function, which is written as, 
\begin{equation}
f(v_i) = w^T \sigma (W^{(1)}v_i + b^{(1)})+b.
\end{equation}
The probability distribution is formalized as
 
\begin{equation}
\label{alpa-add}
\alpha = [f(v_i)]_{i=1}^n,
\end{equation}
\begin{equation}
p(z|\bm{v}) = softmax(\alpha).
\end{equation}
The final output $\bm{s}$ of self-attention is similar to the vanilla attention mechanism above, i.e.,
\begin{equation}
\label{self-attn}
\bm{s}=p(z|\bm{v})\odot \bm{v}
\end{equation}

If the output is used as sequence encoding,  $s$ is the weighted average of sampling a visit according to its importance, i.e.,
\begin{equation}
\label{e_code2visit}
s = \sum_{i=1}^n p(z = i|\bm{v})\odot v_i.
\end{equation}

\subsubsection{Multi-Head Attention}
The multi-head attention mechan-ism\cite{vaswani2017attention} relies on self-attention, which operates on a query $Q$, a key $K$ and a value $V$:

\begin{equation}
\mathrm{Attention}(Q,K,V) =
p(z|Q,K)V = p(z|\bm{v})\bm{v}
\end{equation}
where $Q$, $K$, $V$, $\bm{v}$ is a $m \times d$ matrix, $m$ denotes the number of visits in a patient journey, $d$ denotes the dimension of embedding.
$\bm{v}$ is a sequence of visit embeddings coming from the output of the previous layer. 

For scaled dot-product attention, 
\begin{equation}
\label{alpa_dot}
\alpha = \frac{QK^T}{\sqrt{d}},
\end{equation}
\begin{equation}
p(z|Q,K) = p(z|\bm{v})= \mathrm{softmax}(\alpha).
\end{equation}

The multi-head attention mechanism obtains $h$ (i.e. one per head) different representations of ($Q,K,V$), computes self-attention for each representation, concatenates the results. This can be expressed in the same notation as Eq.(\ref{self-attn}):
\begin{equation}
\label{head}
\mathrm{head}_i = \mathrm{Attention}(QW^Q_i, KW^K_i, VW^V_i)
\end{equation}
\begin{equation}
\label{multi-head}
\mathrm{MultiHead}(Q,K,V) = \mathrm{Concat}(\mathrm{head}_1,...,{head}_h)W^O
\end{equation}
where the projections are parameter matrices $W^Q_i \in \mathbb{R}^{d\times d_k}, W^K_i \in \mathbb{R}^{d\times d_k}, W^V_i \in \mathbb{R}^{d\times d_v}$ and $W^O \in \mathbb{R}^{hd_v\times d}$, $d_k = d_v = d/h$.

\section{Proposed Model}\label{Method}

This section begins with introducing the vanilla and masked encoder, followed by the masked temporal encoder and bidirectional encoder representation network for patient journeys to predict medical outcomes. 

\subsection{Masked Encoder Block}
\begin{figure}[t]
	\centering
	\includegraphics[width=0.3\textwidth]{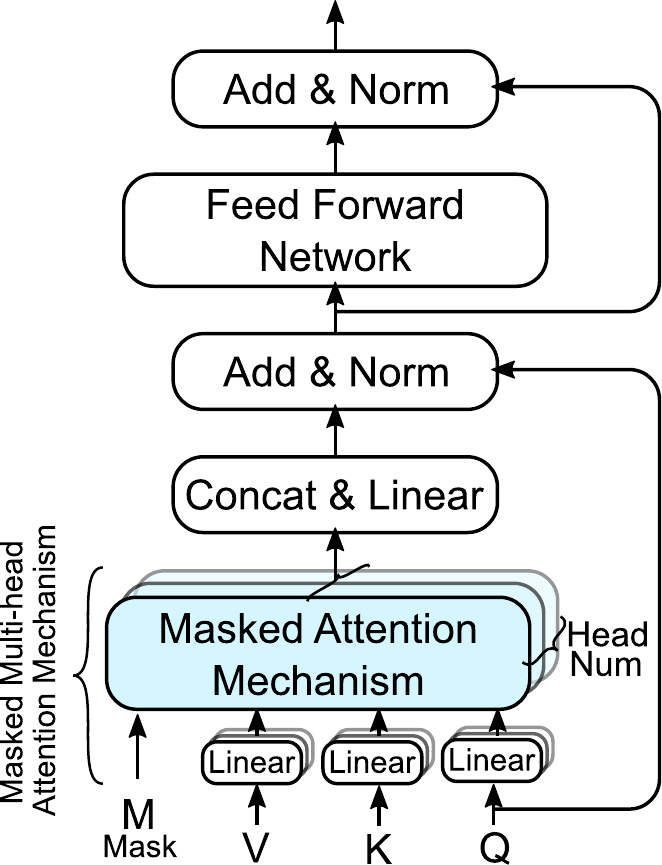}
	\caption{\small Masked encoder (MasEnc) block.} \label{residual_self_attn}
\end{figure}

Because self-attention was originally designed for NLP tasks, it does not consider temporal relationships within inputs. Obviously, this is very important when modeling sequential medical events. Inspired by the work of Vaswani et al.~\cite{vaswani2017attention} with a transformer and Shen et al.~\cite{shen2018disan,shen2018bi} with masked multi-dimensional self-attention, we developed the masked encoder to capture the contextual and temporal relationships between visits. The structure of our masked encoder is shown in Fig.~\ref{residual_self_attn}, which has two sub-layers.  

The first is a multi-head self-attention mechanism, which is rewritten into a temporal-dependent format,
\begin{equation}
\label{self-att}
\mathrm{Attention}(Q,K,V,M) = \mathrm{softmax}(\alpha+M)V,
\end{equation}
where $\alpha$ outlined in Eq.(\ref{alpa-add}) and (\ref{alpa_dot}) work to capture the contextual relationship between the visits, and $M$ to capture the temporal dependency between those visits. There are three types of temporal order information forward, backward and diagonal-disabled mask. There is no temporal order information between medical codes in a visit. In this context, self-attention usually requires disabling the attention of each code to itself~\cite{hu2017reinforced}, which is effectively the same as applying a diagonal-disabled mask such that,
\begin{equation}
\label{diag_mask}
M^{diag}_{ij} = \begin{cases} 0,& i \neq j\\ -\infty, & i = j\end{cases}
\end{equation}

Similarly, masks can be used to encode temporal order information between visits as an attention output. Our approach also incorporates two masks, i.e., a forward mask $M^{fw}$ and backward mask $M^{bw}$, i.e., 
\begin{equation}
\label{f_mask}
M^{fw}_{ij} = \begin{cases} 0,& i < j\\ -\infty, & otherwise\end{cases}
\end{equation}
\begin{equation}
\label{b_mask}
M^{bw}_{ij} = \begin{cases} 0,& i > j\\ -\infty, & otherwise\end{cases}
\end{equation}
The forward mask $M^{fw}$, only relates later visits \textit{j} to earlier visits \textit{i}, and vice versa with the backward one. 

The second sub-layer is a simple, position-wise, fully connected feed-forward network. We employ a residual connection~\cite{he2016deep} around each of the two sub-layers, followed by layer normalization~\cite{ba2016layer}. That is, the output of each layer is LayerNorm($\bm{v}$ + sublayer($\bm{v}$)), where sublayer($\bm{v}$) is the function implemented by the layer itself~\cite{vaswani2017attention}.

\subsection{Bidirectional Temporal Encoder Network}

\begin{figure}[!htb]
	\centering
	\includegraphics[width=0.48\textwidth]{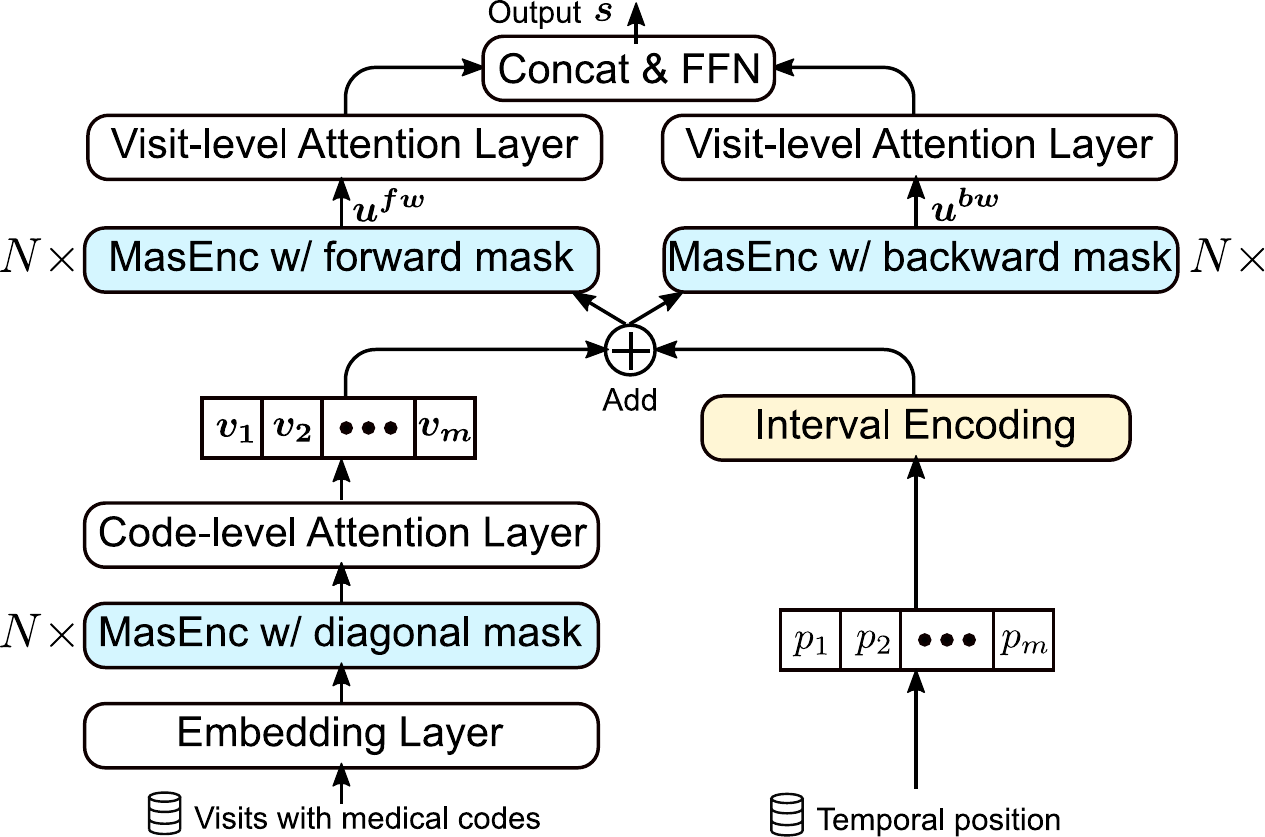}
	\caption{ \small Bidirectional temporal encoder network (BiteNet) for patient journey embedding. }  \label{BiteNet}
\end{figure}

We propose a patient journey embedding model, called \textbf{Bi}directional \textbf{t}emporal \textbf{e}ncoder \textbf{Net}work (BiteNet), with MasEnc as its major components. The architecture of BiteNet is shown in Fig.~\ref{BiteNet}. 
In BiteNet, the embedding layer is applied to the input medical codes of visits, and its output is processed by a stack of $N$ MasEnc blocks with diagonal mask $M^{diag}$ in Eq.(\ref{diag_mask}) and the code-level attention layer to generate visit embeddings $[v_1, v_2, ...,v_m]$ given in Eq.(\ref{e_code2visit}). The visit interval encodings are then added to the visit embeddings, followed by two parameter-untied stacks of $N$ MasEnc blocks with forward mask $M^{fw}$ in Eq.(\ref{f_mask}) and $M^{bw}$ in Eq.(\ref{b_mask}), respectively. Their outputs are denoted by $u^{fw}, u^{bw} \in \mathbb{R}^{d\times m}$. The visit-level attention layers in Eq.(\ref{e_code2visit}) are applied to the outputs followed by the concatenation layer to generate output $u^{bi} \in \mathbb{R}^{2d}$. To complete the process, a feed-forward layer consisting of dense and softmax layers is employed to generate a categorical distribution over targeted categories.

\paragraph{Interval Encoding.} Although MasEnc incorporates information on the order of visits in a patient journey, the relative time intervals between visits is an important factor in longitudinal studies. We include information on the visit intervals $p=[p_{1}, p_{2}, ...,p_{n}]$ in the sequence. In particular, we add interval encodings to the visit embeddings of the sequence. There are two steps to perform interval encoding: 1) building a lookup table with dimensions of $m \times d$, where $m$ is the days of the dataset span (e.g., $m=11 \times 365$ over 11 years), and $d$ is the dimension of code embedding; 2) updating the lookup table during training.
The \textit{d}-dimensional interval encoding is then added to the visit embedding with the same dimension. 

\section{Experiments}\label{Experim}

We conducted experiments based on the MIMIC-III dataset to compare the performance of our proposed method, BiteNet, with several state-of-the-art methods. We also evaluate the influence of the model components via ablation studies. Building on this, we visualize the importance of visits for patients and diagnoses in visits derived from the visit-level and code-level attention weightings, respectively. At last, we compare the performance between BiteNet and baseline models on unsupervised tasks. The source code is available in https://github.com/Xueping/BiteNet.

\subsection{Data Description}
\subsubsection{Dataset} MIMIC-III~\cite{Johnson_2016} is an open-source, large-scale, de-identified dataset of ICU patients and their EHRs. The dataset consists of 46k+ ICU patients with 58k+ visits collected over 11 years. In this paper, we consider two sub-datasets: Dx and Dx\&Tx, where Dx is a dataset which only includes diagnosis codes for each visit, and Dx\&Tx is another dataset which includes diagnosis and procedure codes for each visit.

\subsubsection{Data Pre-processing} We chose patients who made at least two visits. All  infrequent diagnoses codes were removed and the threshold was empirically set to 5. In summary, we extract 7,499 patients, each with an average of 2.66 visits; the average number of diagnoses and procedures in each visit are 13 and 4, respectively.

\subsection{Experiment Setup}
\subsubsection{Supervised Prediction Tasks}

The two tasks we selected to evaluate the performance of our proposed model are to predict re-admission~\cite{xiao2018readmission} and future diagnosis~\cite{Rajkomar_Google_2018}. 
\begin{itemize}
	\item \textbf{Re-admission (Readm)} is a standard measure of the quality of care. We predicted unplanned admissions within 30 days following discharge from an indexed visit. A visit is considered a “re-admission” if its admission date is within 30 days of discharge from an eligible indexed visit~\cite{Rajkomar_Google_2018}.
	\item \textbf{Diagnoses} reflect the model's understanding of a patient's problems. In the experiments, we aim to predict diagnosis categories instead of the real diagnosis codes, which are the nodes in the second hierarchy of the ICD9 codes\footnote{http://www.icd9data.com}.
\end{itemize}

\subsubsection{Unsupervised Tasks}
Two tasks of \textbf{clustering} and \textbf{nearest neighbor search (NNS)}~\cite{MCE_Cai_2018} in unsupervised learning were conducted to evaluate the quality of the medical code embedding results. We selected the ground truth by using a well-organized ontology, Clinical Classifications Software (CCS)\footnote{https://www.hcup-us.ahrq.gov}. CCS provides a way to classify diagnoses and procedures into a limited number of categories by aggregating individual ICD9 codes into broad diagnosis and procedure groups to facilitate statistical analysis and reporting\footnote{ https://www.hcup-us.ahrq.gov/toolssoftware/ccs/CCSUsersGuide.pdf}. CCS aggregates ICD9 diagnosis codes into 285 mutually exclusive categories. We obtained 265 categories and 61,630 near neighbor pairs for clustering and the nearest neighbor search, respectively.

\renewcommand{\arraystretch}{1.4}
\begin{table*}[t]
	\caption{ Prediction performance comparison of future re-admission and diagnoses (Dx is for diagnosis, and Tx is for procedure).}\label{tab_tab2}
	\centering
	\scalebox{1}{%
		\begin{tabular}{|l|l|c|c|c|c|c|c|c|}
			\hline
			\textbf{Data}&\textbf{Model}&\multicolumn{1}{|c}{\textbf{Readm}}&\multicolumn{6}{|c|}{\textbf{Diagnosis Precision@k}} \\
			
			\cline{4-9} & & \textbf{{\small (PR-AUC)}} &\textbf{k=5} & \textbf{k=10} &\textbf{k=15}& \textbf{k=20} &\textbf{k=25}& \textbf{k=30} \\ \hline
			&RNN     & 0.3021$\pm$0.0051 & 0.6330$\pm$0.0055& 0.5874$\pm$0.0035 &0.6309$\pm$0.0047 & 0.6977$\pm$0.0064 &0.7601$\pm$0.0053 & 0.8068$\pm$0.0011 \\ \cline{2-9}
			&BRNN    & 0.3119$\pm$0.0078 & 0.6362$\pm$0.0051 & 0.5925$\pm$0.0031 &0.6365$\pm$0.0026 & 0.7014$\pm$0.0035 &0.7596$\pm$0.0049 & 0.8128$\pm$0.0026 \\ \cline{2-9}
			&RETAIN  & 0.3014$\pm$0.0127 & 0.6498$\pm$0.0075 & 0.5948$\pm$0.0055 &0.6368$\pm$0.0054 & 0.6999$\pm$0.0060 &0.7627$\pm$0.0039& 0.8102$\pm$0.0042 \\ \cline{2-9}
			Dx&Deepr   & 0.2999$\pm$0.0190 & 0.6434$\pm$0.0030 & 0.5865$\pm$0.0025 &0.6257$\pm$0.0033& 0.6981$\pm$0.0018 &0.7604$\pm$0.0044& 0.8113$\pm$0.0047 \\ \cline{2-9}
			&Dipole& 0.2841$\pm$0.0128 & 0.6484$\pm$0.0062 & 0.5997$\pm$0.0042 &0.6339$\pm$0.0041& 0.7034$\pm$0.0026 &0.7620$\pm$0.0024& 0.8121$\pm$0.0012 \\ \cline{2-9}
			&SAnD    & 0.2979$\pm$0.0263 & 0.6179$\pm$0.0147 & 0.5709$\pm$0.0136 & 0.6100$\pm$0.0126& 0.6805$\pm$0.0123 &0.7464$\pm$0.0119& 0.7959$\pm$0.0109 \\ \cline{2-9}
			&BiteNet & \textbf{0.3266$\pm$0.0047} & \textbf{0.6615$\pm$0.0024} & \textbf{0.6019$\pm$0.0056} &\textbf{0.6432$\pm$0.0031}& \textbf{0.7104$\pm$0.0031 } &\textbf{0.7757$\pm$0.0046}& \textbf{0.8245$\pm$0.0053} \\ \hline
			&RNN        & 0.3216$\pm$0.0047 & 0.6317$\pm$0.0055 & 0.5857$\pm$0.0029 &0.6310$\pm$0.0037 & 0.6973$\pm$0.0015 &0.7616$\pm$0.0021&0.8093$\pm$0.0019 \\ \cline{2-9}
			&BRNN       & 0.3270$\pm$0.0065 & 0.6402$\pm$0.0064 & 0.5961$\pm$0.0025 &0.6386$\pm$0.0032 & 0.7088$\pm$0.0036 &0.7662$\pm$0.0018& 0.8138$\pm$0.0018 \\ \cline{2-9}
			&RETAIN     & 0.3161$\pm$0.0107 & 0.6552$\pm$0.0058 & 0.6048$\pm$0.0047 & 0.6429$\pm$0.0060 & 0.7085$\pm$0.0042 & 0.7675$\pm$0.0042 & 0.8166$\pm$0.0039 \\ \cline{2-9}
			Dx\&Tx&Deepr      & 0.3142$\pm$0.0150 & 0.6391$\pm$0.0038 & 0.5947$\pm$0.0040 &0.6330$\pm$0.0042& 0.7004$\pm$0.0043 &0.7633$\pm$0.0041& 0.8125$\pm$0.0031 \\ \cline{2-9}
			&Dipole & 0.2899$\pm$0.0093 & 0.6515$\pm$0.0078 & 0.6097$\pm$0.0050 &0.6415$\pm$0.0044& 0.7121$\pm$0.0064 &0.7691$\pm$0.0065& 0.8149$\pm$0.0046 \\ \cline{2-9}
			&SAnD     & 0.2996$\pm$0.0135 & 0.6242$\pm$0.0104 & 0.5774$\pm$0.0106 &0.6233$\pm$0.0092& 0.6878$\pm$0.0089 &0.7510$\pm$0.0107& 0.8004$\pm$0.0122 \\ \cline{2-9}
			&BiteNet  & \textbf{0.3357$\pm$0.0045} & \textbf{0.6705$\pm$0.0045}&\textbf{0.6117$\pm$0.0033}& \textbf{0.6511$\pm$0.0035}& \textbf{0.7187$\pm$0.0046}& \textbf{0.7799$\pm$0.0067}& \textbf{0.8289$\pm$0.0068} \\ \hline
		\end{tabular}
	}
\end{table*}

\subsubsection{Baseline Methods}
We compare the performance of our proposed model against several baseline models:

\begin{itemize}
\item \textbf{RNN} and \textbf{BRNN}, we directly embed visit information into the vector representation $v_t$ by summation of embedded medical codes in the visit, and then feed this embedding to the GRU. The hidden state $h_t$ produced by the GRU is used to predict the (\textit{t}+1)-th visit information.
\item \textbf{RETAIN}~\cite{choi2016retain}: which learns the medical concept embeddings and performs heart failure prediction via the reversed RNN with the attention mechanism.
\item \textbf{Deepr}~\cite{nguyen2016deepr}: which is a multilayered architecture based on convolutional neural networks (CNNs) that learn to extract features from medical records and predict future risk.
\item \textbf{Dipole}~\cite{ma2017dipole}, which uses bidirectional recurrent neural networks and three attention mechanisms (location-based, general, concatenation-based) to predict patient visit information. We chose concatenation-based Dipole as a baseline method.
 \item \textbf{SAnD}~\cite{song2018attend}: which employs a masked, self-attention mechanism, and uses positional encoding and dense interpolation strategies to incorporate temporal order to generate a sequence-level prediction.
\end{itemize}

\begin{figure*}[t]
	\centering
	\includegraphics[width=1\textwidth]{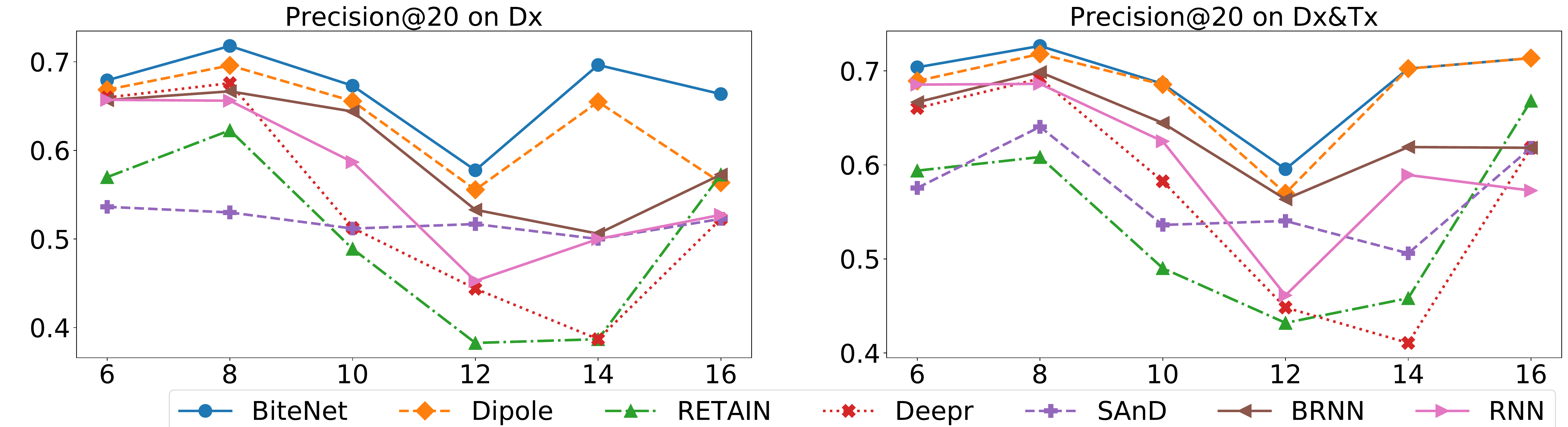}
	\caption{\small Robustness comparison regarding Precision @20 on two datasets. Length of the visit sequence varies from 6 to 16.} \label{fig_robustness}
\end{figure*}

\subsubsection{Evaluation metrics} 
The two evaluation metrics used are:
\begin{itemize}
\item \textbf{PR-AUC:}
(Area under Precision-Recall Curve), to evaluate the likelihood of re-admission. PR-AUC is considered to be a better measure for imbalanced data like ours. 
\item \textbf{precision@k:}
which is defined as the correct medical codes in top k divided by $\mathrm{min}(k, |y_t|)$, where $|y_t|$ is the number of category labels in the (\textit{t}+1)-th visit~\cite{Ma2018-gu}. We report the average values of precision@k in the experiments. We vary \textit{k} from 5 to 30. The greater the value, the better the performance.

\item \textbf{Accuracy@k:}
to evaluate the performance of the nearest neighbor search, which is defined as the number of the nearest neighbors in top \textit{k} divided by \textit{k}. We report the average values of Accuracy@k in the experiments. We vary \textit{k} from 1 to 10. The greater the value, the better the performance.

\item \textbf{NMI:}
normalized mutual information for clustering performance.
\end{itemize}

\begin{figure*}[!htb]
	\centering
	\includegraphics[width=1\textwidth]{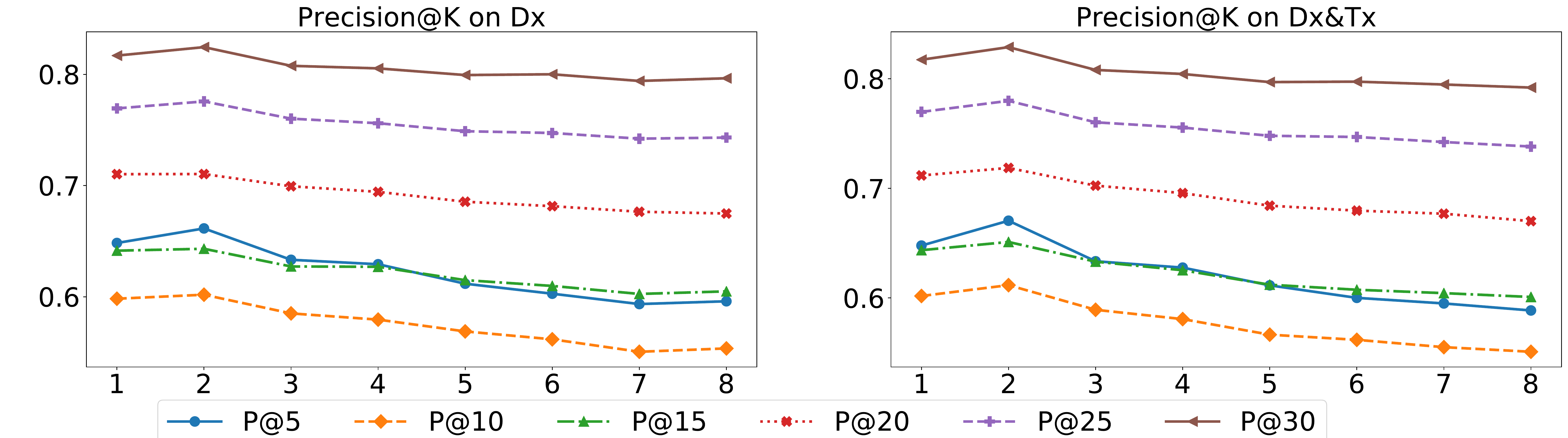}
	\caption{\small Diagnoses Precision@\{5,10,15,20,25,30\} on two datasets. The MasEnc layer number $N$ varies from 1 to 8.} \label{fig_precision_layers}
\end{figure*}

\begin{figure*}[!htb]
	\centering
	\includegraphics[width=1\textwidth]{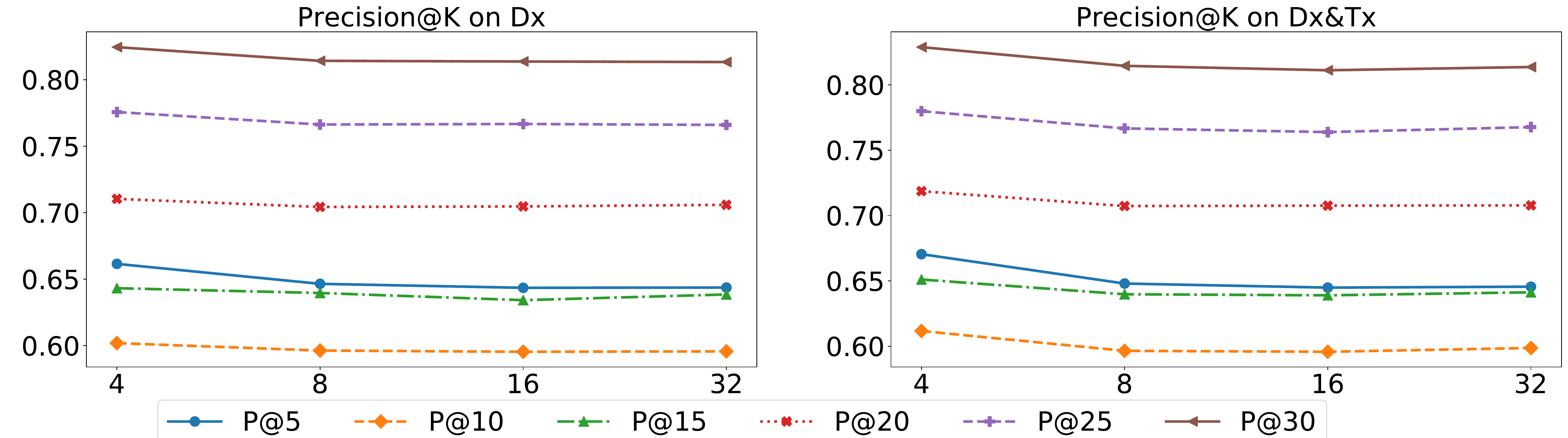}
	\caption{\small Diagnoses Precision@\{5,10,15,20,25,30\} on two datasets when $N=2$. The number of heads in multi-head attention varies from 4 to 32.} \label{fig_precision_heads}
\end{figure*}

\subsubsection{Implementation}
We implement all the approaches with Tensorflow 2.0. For training models, we use RMSprop with a minibatch of 32 patients and 10 epochs. The drop-out rate is 0.1 for all approaches. The data split ratio is 0.8:0.1:0.1 for training, validation and testing sets. In order to fairly compare the performance, we set the same embedding dimension \textit{d} = 128 for all the baselines and the proposed model. 

\subsection{Results of Prediction Tasks}

\subsubsection{Prediction Performance Comparison}
Table~\ref{tab_tab2} reports the results of the two prediction tasks on the two datasets - future re-admissions and diagnoses. The results show that BiteNet outperforms all the baselines. This demonstrates that the superiority of our framework results from the explicit consideration of the inherent hierarchy of EHRs and the contextual and temporal dependencies which are incorporated into the representations. In addition, we note that the performance obtained by the models remains approximately the same (RNN), or even drops by up to 0.43\% (Deepr), after adding the procedure to the training data over Precision@5. The underlying reason might be that the future diagnosis prediction is less sensitive to the procedures, with the result that the relationships among them cannot be well captured when predicting few diagnoses. Still, by using attention mechanisms, Dipole, SAnD and our BiteNet model, we achieve a marginal improvement when comparing the performance of using Dx\&Tx for training to that of Dx. This implies that attention could play an important role in the learned representations of the procedures. For the performance of future re-admission, we note that all models achieve an improvement when adding procedure information in the training stage, indicating that the additional procedures benefit for the future re-admission prediction.

\subsubsection{Robustness for the lengthy sequence of visits}

We conducted a set of experiments to evaluate the robustness of BiteNet by varying the length of sequential visits of a patient from 6 to 16, considered to be long patient journeys in the medical domain.  Fig.~\ref{fig_robustness} shows the performance of future diagnoses prediction on Precision@20. Overall, the model BiteNet outperforms the other models as the length of a patient's sequential visits increases diagnoses prediction over the dataset with only diagnoses information. Our BiteNet model achieves the best performance when the lengths of sequential visits are 6, 8 and 12 over Dx\&Tx. In particular, the performances of Dipole and BRNN are comparative with our BiteNet and follow a similar trend  with an increase in the length of sequential visits. From these results, a conclusion can be drawn that the temporal order masks in the MasEnc module and interval encoding in our framework play a vital role in capturing the lengthy temporal dependencies between patients' sequential visits.

\begin{table*}[t]
	\centering
	\caption{\small Ablation Performance Comparison.}
	\scalebox{1}{%
		\begin{tabular}{|l|l|c|c|c|c|c|c|c|}
			\hline
			\textbf{Data}&\textbf{Ablation}&\multicolumn{1}{c}{\textbf{Readm}}&\multicolumn{6}{|c|}{\textbf{Diagnosis Precision@k}} \\
			
			\cline{4-9} & & \textbf{{\small (PR-AUC)}} &\textbf{k=5} & \textbf{k=10} &\textbf{k=15}& \textbf{k=20} &\textbf{k=25}& \textbf{k=30} \\
			\hline
			&Attention  & 0.3015$\pm$0.0068 & 0.6506$\pm$0.0050 & 0.6014$\pm$0.0028 &0.6440$\pm$0.0032 & 0.7125$\pm$0.0022 &0.7712$\pm$0.0020& 0.8174$\pm$0.0023 \\ \cline{2-9}
			Dx&DireMask    & 0.3026$\pm$0.0132 & 0.6528$\pm$0.0051 & 0.6047$\pm$0.0031 &0.6460$\pm$0.0029& 0.7128$\pm$0.0035 &0.7711$\pm$0.0030& 0.8191$\pm$0.0034\\ \cline{2-9}
			&Interval & 0.2921$\pm$0.0111 & 0.6469$\pm$0.0041 & 0.5947$\pm$0.0050 &0.6361$\pm$0.0029& 0.7054$\pm$0.0025 &0.7667$\pm$0.0028& 0.8139$\pm$0.0020\\ \cline{2-9}
			&BiteNet & \textbf{0.3266$\pm$0.0047} & \textbf{0.6615$\pm$0.0124} & \textbf{0.6019$\pm$0.0056} &\textbf{0.6432$\pm$0.0031}& \textbf{0.7104$\pm$0.0031 } &\textbf{0.7757$\pm$0.0046}& \textbf{0.8245$\pm$0.0053} \\ \hline
			&Attention  &0.3017$\pm$0.0070 & 0.6453$\pm$0.0046 & 0.5984$\pm$0.0030 &0.6424$\pm$0.0027 & 0.7092$\pm$0.0036 &0.7701$\pm$0.0020& 0.8176$\pm$0.0009 \\ \cline{2-9}
			Dx\&&DireMask    & 0.3046$\pm$0.0124 & 0.6487$\pm$0.0050 & 0.6015$\pm$0.0021 &0.6437$\pm$0.0023& 0.7117$\pm$0.0009 &0.7728$\pm$0.0013& 0.8193$\pm$0.0008\\ \cline{2-9}
			Tx &Interval & 0.3109$\pm$0.0109 & 6420$\pm$0.0018 & 0.5945$\pm$0.0027 &0.6352$\pm$0.0034& 0.7060$\pm$0.0021 &0.7664$\pm$0.0013& 0.8148$\pm$0.0016\\ \cline{2-9}
			&BiteNet  & \textbf{0.3357$\pm$0.0045} & \textbf{0.6705$\pm$0.0045}&\textbf{0.6117$\pm$0.0033}& \textbf{0.6511$\pm$0.0035}& \textbf{0.7187$\pm$0.0046}& \textbf{0.7799$\pm$0.0067}& \textbf{0.8289$\pm$0.0068} \\ \hline
		\end{tabular}
	}
	\label{tab_ablation}
\end{table*}

\subsubsection{Effectiveness of varying stack \textit{N} of MasEnc layers}

In this set of experiments, we evaluated the performance of our proposed model BiteNet by varying the stack $N$ of MasEnc layers from 1 to 8, where the number of heads in multi-head attention is set to 4. Fig.~\ref{fig_precision_layers} shows the results. Overall, BiteNet achieved the best performance of future diagnoses prediction when the layer number of MasEnc block is 2. After this, the performance was slightly declining as the number of layers increased. However, the precision of future diagnoses prediction dropped to a minimum when $N=7$ and then slightly increased when $N=8$ over dataset Dx. Overall, we find the performance of Precision@30 outperformed the others, and Precision@10 was the worth over both Dx and Dx\&Tx. 

\subsubsection{Effectiveness of varying heads \textit{h} of multi-head attention}

In this set of experiments, we varied the number of heads $h$ in multi-head attention from 4 to 32 to evaluate the effectiveness of the proposed BiteNet, where the stack $N$ of MasEnc layers is set to 2.  From Fig.~\ref{fig_precision_heads}, we can observe that the performance of future diagnoses prediction drops when the number of heads becomes larger.  Particularly, the prediction precision of BiteNet remains stable or drops slightly over both Dx and Dx\&Tx. It demonstrates that our proposed model, BiteNet, is less sensitive to the number of heads in multi-head attention.

\subsubsection{Ablation Study}

We performed a detailed ablation study to examine the contributions of each of the model's components to the prediction tasks. There are three components: (Attention) the two attention pooling layers to learn the visits from the embedded medical codes and learn the patient journey from the embedded visits; (DireMask) the direction mask in MasEnc module, and (Interval) the interval encodings to be added to the learned visit embeddings. 
\begin{itemize}
	\item  \textbf{Attention:} replace each of the two attention pooling layers with a simple summation layer; 
	\item  \textbf{DireMask:} remove the diagonal-disabled, forward and backward directional mask in the MasEnc module; 
	\item  \textbf{Interval:} remove the interval encodings module; 
	\item  \textbf{BiteNet:} our model as proposed in the paper.
\end{itemize}	

From Table~\ref{tab_ablation}, we see that the full complement of BiteNet achieves superior accuracy over the ablated models. Specifically, we note that Interval and Attention contribute the highest accuracy to re-admission prediction over Dx and Dx\&Tx, respectively. It gives us confidence in using the Interval encoding and Attention pooling to learn the patient journey representations without sufficient data. Significantly, it is clear that the MasEnc component provides valuable information with the learned embeddings of the patient journey for the performance of diagnoses prediction over Dx\&Tx. Specifically, BiteNet predicted re-admissions with 3.45\% more accuracy on Dx and with 3.40\% over Dx\&Tx. Similarly, BiteNet predicted diagnoses precision@5 with 1.46\% more accurately than Dx and diagnoses precision@30 with 1.41\% more accuracy than Dx\&Tx.

\begin{table*}[t]
	\caption{ Prediction performance comparison of future re-admission and diagnoses (Dx is for diagnosis, and Tx is for procedure).}\label{tab_unsuper}
	\centering
	\scalebox{1}{%
		\begin{tabular}{|l|l|c|c|c|c|}
			\hline
			\textbf{Data}&\textbf{Model}&\multicolumn{1}{c}{\textbf{Clustering}}&\multicolumn{3}{|c|}{\textbf{NNS Accuracy@k}} \\
			
			\cline{4-6} & & \textbf{{\small (NMI)}} &\textbf{k=1} & \textbf{k=5} &\textbf{k=10} \\ \hline
			&RNN     & 0.4824$\pm$0.0054 & 0.1462$\pm$0.0133 & 0.1236$\pm$0.0043 &0.1154$\pm$0.0030\\ \cline{2-6}
			&BRNN    & 0.4842$\pm$0.0035 & 0.1524$\pm$0.0078 & 0.1204$\pm$0.0053 &0.1084$\pm$0.0062 \\ \cline{2-6}
			&RETAIN  & 0.4844$\pm$0.0084 & 0.1556$\pm$0.0154 & 0.1254$\pm$0.0041 
			&0.1100$\pm$0.0030 \\ \cline{2-6}
			Dx&Deepr & 0.4571$\pm$0.0059 & 0.1120$\pm$0.0112 & 0.0936$\pm$0.0040 &0.0914$\pm$0.0041\\ \cline{2-6}
			&Dipole  & 0.4822$\pm$0.0046 & 0.1504$\pm$0.0117 & 0.1156$\pm$0.0038 &0.1084$\pm$0.0030\\ \cline{2-6}
			&SAnD    & 0.4717$\pm$0.0044 & 0.1243$\pm$0.0096 & 0.1148$\pm$0.0071 &0.0984$\pm$0.0041\\ \cline{2-6}
			&BiteNet & \textbf{0.4856$\pm$0.0035} & \textbf{0.1584$\pm$0.0075} & \textbf{0.1284$\pm$0.0035} &\textbf{0.1154$\pm$0.0028} \\ \hline
			&RNN        & 0.4838$\pm$0.0036 & 0.1505$\pm$0.0118 & 0.1236$\pm$0.0032 &0.1148$\pm$0.0028\\ \cline{2-6}
			&BRNN       & 0.4851$\pm$0.0028 & 0.1643$\pm$0.0069 & 0.1240$\pm$0.0036 &0.1150$\pm$0.0048\\ \cline{2-6}
			&RETAIN     & 0.4867$\pm$0.0078 & 0.1744$\pm$0.0153 & 0.1312$\pm$0.0032 &0.1142$\pm$0.0036\\ \cline{2-6}
			Dx\&Tx&Deepr      & 0.4658$\pm$0.0051 & 0.1283$\pm$0.0100 & 0.1116$\pm$0.0028 &0.1010$\pm$0.0006\\ \cline{2-6}
			&Dipole     & 0.4858$\pm$0.0056 & 0.1744$\pm$0.0110 &0.1343$\pm$0.0040& 0.1188$\pm$0.0042\\ \cline{2-6}
			&SAnD       & 0.4722$\pm$0.0040 & 0.1264$\pm$0.0095 & 0.1084$\pm$0.0045 &0.1014$\pm$0.0047\\ \cline{2-6} 
			&BiteNet  & \textbf{0.4873$\pm$0.0034} & \textbf{0.1758$\pm$0.0061} & \textbf{0.1492$\pm$0.0030} &\textbf{0.1244$\pm$0.0024}\\ \hline
			
		\end{tabular}
	}
	
\end{table*}
\subsubsection{Visualization and Explainability}

\begin{figure}[t]
	\centering
	\includegraphics[width=0.48\textwidth]{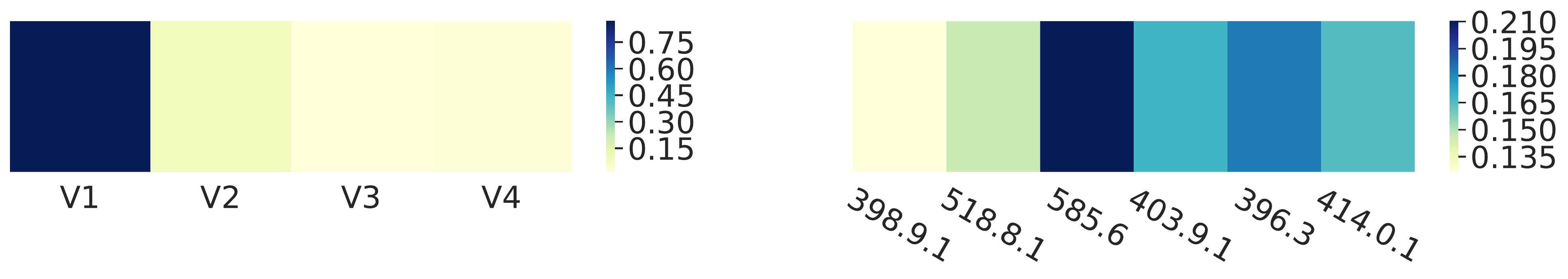}
	\caption{\small Importance of visits for patient 1, and importance of diagnoses in visit 1.} \label{fig_importance}
\end{figure}

\begin{figure}[t]
	\centering
	\includegraphics[width=0.48\textwidth]{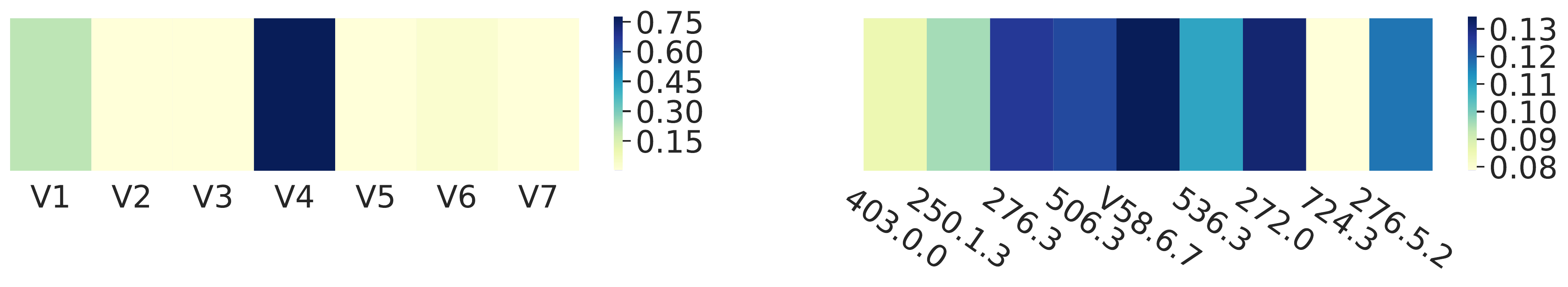}
	\caption{\small Importance of visits for patient 2, and importance of diagnoses in visit 4.} \label{fig_importance2}
\end{figure}

One aspect of this method is that it hierarchically compresses medical codes into visits and visits into patient journeys. At each level of aggregation, the model decides the importance of the lower-level entries on the upper-level entry, which makes the model explainable. To showcase this feature, we visualize two patient journeys. These examples come from the re-admission prediction task with the MIMIC dataset. From the importance distribution of the patient visits, we analyze the most essential visits to future re-admission. For example, Visit 1 in Fig.~\ref{fig_importance} was the most influential factor in Patient 1's re-admission. After closely examining Visit 1, we found a vital insight into Diagnosis 3 (ICD 585.6), i.e., end-stage renal disease, which would obviously cause frequent and repeated re-admissions to the hospital. in Fig.~\ref{fig_importance2}, Patient 2 had 7 visits. Visit 4 contributed most to the re-admissions. Again, the diagnoses reveal the cause: long-term use of insulin (ICD V58.6.7) and pure hypercholesterolemia (ICD 272.0).


\subsubsection{Results of Unsupervised Tasks}
We use the clustering and nearest neighbor search tasks to evaluate the embedding results on two datasets: Dx and Dx\&Tx. We choose K-Means as the clustering algorithm, and NMI to evaluate the learned representations for the diagnosis codes. The embedding of diagnosis codes is derived from the proposed model BiteNet on the task of future diagnoses prediction. 

NMI for clustering and Accuracy@k ($k=1,5,10$) for NNS are reported in Table~\ref{tab_unsuper}, where we highlight the best results. Overall, the model BiteNet outperforms the other baseline models on the two unsupervised tasks over two datasets. From the table, we find that the RNN-based models (RNN, BRNN, RETAIN and Dipole) obtained better performance in medical code embedding compared to the CNN-based models (Deepr and SAnD). This demonstrates that temporal dependency on EHR data is significantly hidden information when embedding medical codes and RNN-based models can capture the temporal relationship. However, our proposed model, BiteNet, outperforms the baseline models including RNN-based, CNN-based, and attention integrated models, because BiteNet captures the contextual information and long-term dependencies between patients' visits, and considers the hierarchical structure of EHR data as well. The superior performance of BiteNet over the other models can also be explained by the introduction of the temporal self-attention model and the incorporation of the contextual information and the temporal interval from the data, which create a better learner of the medical code embeddings. 

\section{Conclusion}\label{Conc}
In this paper, we proposed a novel prediction model called \textbf{BiteNet}. The model framework comprises a MasEnc module that captures the contextual information and the temporal relationships between the visits in a patient's healthcare journey and attention pooling that construct the hierarchical structure of three-levelled EHR data. The output is a representation of a patient journey that, once learned by the model, can be used to predict medical outcomes with an end-to-end sole self-attention network. We evaluated BiteNet's performance of the model against several baseline methods with supervised and unsupervised tasks, and conducted an ablation study to examine the contributions of each component. The results show that BiteNet produces more accurate predictions than baseline methods.

\section{Acknowledgments}
This work was supported in part by the Australian Research Council (ARC) under Grant LP160100630, LP180100654 and DE190100626.

\bibliographystyle{IEEEtran}
\bibliography{References.bib}

\begin{thebibliography}{10}
\providecommand{\url}[1]{#1}
\csname url@samestyle\endcsname
\providecommand{\newblock}{\relax}
\providecommand{\bibinfo}[2]{#2}
\providecommand{\BIBentrySTDinterwordspacing}{\spaceskip=0pt\relax}
\providecommand{\BIBentryALTinterwordstretchfactor}{4}
\providecommand{\BIBentryALTinterwordspacing}{\spaceskip=\fontdimen2\font plus
\BIBentryALTinterwordstretchfactor\fontdimen3\font minus
  \fontdimen4\font\relax}
\providecommand{\BIBforeignlanguage}[2]{{%
\expandafter\ifx\csname l@#1\endcsname\relax
\typeout{** WARNING: IEEEtran.bst: No hyphenation pattern has been}%
\typeout{** loaded for the language `#1'. Using the pattern for}%
\typeout{** the default language instead.}%
\else
\language=\csname l@#1\endcsname
\fi
#2}}
\providecommand{\BIBdecl}{\relax}
\BIBdecl

\bibitem{Shickel_2018}
B.~Shickel, P.~J. Tighe, A.~Bihorac, and P.~Rashidi, ``Deep ehr: a survey of
  recent advances in deep learning techniques for electronic health record
  (ehr) analysis,'' \emph{IEEE J Biomed Health Inform}, vol.~22, no.~5, pp.
  1589--1604, 2018.

\bibitem{Rajkomar_Google_2018}
A.~Rajkomar, E.~Oren, K.~Chen, A.~M. Dai, N.~Hajaj, M.~Hardt, P.~J. Liu,
  X.~Liu, J.~Marcus, M.~Sun \emph{et~al.}, ``Scalable and accurate deep
  learning with electronic health records,'' \emph{npj Digital Medicine},
  vol.~1, no.~1, p.~18, 2018.

\bibitem{Qiao_2018}
Z.~Qiao, S.~Zhao, C.~Xiao, X.~Li, Y.~Qin, and F.~Wang, ``Pairwise-ranking based
  collaborative recurrent neural networks for clinical event prediction.'' in
  \emph{IJCAI}, 2018, pp. 3520--3526.

\bibitem{Choi_2016}
E.~Choi, M.~T. Bahadori, E.~Searles, C.~Coffey, M.~Thompson, J.~Bost,
  J.~Tejedor-Sojo, and J.~Sun, ``Multi-layer representation learning for
  medical concepts,'' in \emph{SIGKDD}.\hskip 1em plus 0.5em minus 0.4em\relax
  ACM, 2016, pp. 1495--1504.

\bibitem{Choi_Bahadori_2017}
E.~Choi, M.~T. Bahadori, L.~Song, W.~F. Stewart, and J.~Sun, ``Gram:
  graph-based attention model for healthcare representation learning,'' in
  \emph{SIGKDD}.\hskip 1em plus 0.5em minus 0.4em\relax ACM, 2017, pp.
  787--795.

\bibitem{choi2018mime}
E.~Choi, C.~Xiao, W.~Stewart, and J.~Sun, ``Mime: Multilevel medical embedding
  of electronic health records for predictive healthcare,'' in \emph{NeurIPS},
  2018, pp. 4552--4562.

\bibitem{ma2017dipole}
F.~Ma, R.~Chitta, J.~Zhou, Q.~You, T.~Sun, and J.~Gao, ``Dipole: Diagnosis
  prediction in healthcare via attention-based bidirectional recurrent neural
  networks,'' in \emph{SIGKDD}.\hskip 1em plus 0.5em minus 0.4em\relax ACM,
  2017, pp. 1903--1911.

\bibitem{choi2016retain}
E.~Choi, M.~T. Bahadori, J.~Sun, J.~Kulas, A.~Schuetz, and W.~Stewart,
  ``Retain: An interpretable predictive model for healthcare using reverse time
  attention mechanism,'' in \emph{NeurIPS}, 2016, pp. 3504--3512.

\bibitem{hochreiter1997long}
S.~Hochreiter and J.~Schmidhuber, ``Long short-term memory,'' \emph{Neural
  computation}, vol.~9, no.~8, pp. 1735--1780, 1997.

\bibitem{cho2014learning}
K.~Cho, B.~Van~Merri{\"e}nboer, C.~Gulcehre, D.~Bahdanau, F.~Bougares,
  H.~Schwenk, and Y.~Bengio, ``Learning phrase representations using rnn
  encoder-decoder for statistical machine translation,''
  \emph{arXiv:1406.1078}, 2014.

\bibitem{song2018attend}
H.~Song, D.~Rajan, J.~J. Thiagarajan, and A.~Spanias, ``Attend and diagnose:
  Clinical time series analysis using attention models,'' in \emph{AAAI}, 2018.

\bibitem{Johnson_2016}
A.~E. Johnson, T.~J. Pollard, L.~Shen, H.~L. Li-wei, M.~Feng, M.~Ghassemi,
  B.~Moody, P.~Szolovits, L.~A. Celi, and R.~G. Mark, ``Mimic-iii, a freely
  accessible critical care database,'' \emph{Scientific data}, vol.~3, p.
  160035, 2016.

\bibitem{vaswani2017attention}
A.~Vaswani, N.~Shazeer, N.~Parmar, J.~Uszkoreit, L.~Jones, A.~N. Gomez,
  {\L}.~Kaiser, and I.~Polosukhin, ``Attention is all you need,'' in
  \emph{NeurIPS}, 2017, pp. 5998--6008.

\bibitem{Bahdanau_2014}
D.~Bahdanau, K.~Cho, and Y.~Bengio, ``Neural machine translation by jointly
  learning to align and translate,'' \emph{arXiv:1409.0473}, 2014.

\bibitem{niu2020multichannel}
K.~Niu, J.~Guo, Y.~Pan, X.~Gao, X.~Peng, N.~Li, and H.~Li, ``Multichannel deep
  attention neural networks for the classification of autism spectrum disorder
  using neuroimaging and personal characteristic data,'' \emph{Complexity},
  vol. 2020.

\bibitem{shen2018disan}
T.~Shen, T.~Zhou, G.~Long, J.~Jiang, S.~Pan, and C.~Zhang, ``Disan: Directional
  self-attention network for rnn/cnn-free language understanding,'' in
  \emph{AAAI}, 2018.

\bibitem{Kim2014-pk}
Y.~Kim, ``Convolutional neural networks for sentence classification,'' in
  \emph{EMNLP}, Oct. 2014, pp. 1746--1751.

\bibitem{jha2018interpretable}
K.~Jha, Y.~Wang, G.~Xun, and A.~Zhang, ``Interpretable word embeddings for
  medical domain,'' in \emph{ICDM}.\hskip 1em plus 0.5em minus 0.4em\relax
  IEEE, 2018, pp. 1061--1066.

\bibitem{Mikolov_2013}
T.~Mikolov, I.~Sutskever, K.~Chen, G.~S. Corrado, and J.~Dean, ``Distributed
  representations of words and phrases and their compositionality,'' in
  \emph{NeurIPS}, 2013, pp. 3111--3119.

\bibitem{Mikolov_2013_b}
T.~Mikolov, K.~Chen, G.~Corrado, and J.~Dean, ``Efficient estimation of word
  representations in vector space,'' \emph{arXiv:1301.3781}, 2013.

\bibitem{peng2019attentive}
X.~Peng, G.~Long, S.~Pan, J.~Jiang, and Z.~Niu, ``Attentive dual embedding for
  understanding medical concepts in electronic health records,'' in
  \emph{IJCNN}.\hskip 1em plus 0.5em minus 0.4em\relax IEEE, 2019, pp. 1--8.

\bibitem{Minarro_2014}
J.~A. Minarro-Gim{\'e}nez, O.~Marin-Alonso, and M.~Samwald, ``Exploring the
  application of deep learning techniques on medical text corpora.'' \emph{Stud
  Health Technol Inform}, vol. 205, pp. 584--588, 2014.

\bibitem{De_Vine_2014}
L.~De~Vine, G.~Zuccon, B.~Koopman, L.~Sitbon, and P.~Bruza, ``Medical semantic
  similarity with a neural language model,'' in \emph{CIKM}.\hskip 1em plus
  0.5em minus 0.4em\relax ACM, 2014, pp. 1819--1822.

\bibitem{Choi_AMIA_2016}
Y.~Choi, C.~Y.-I. Chiu, and D.~Sontag, ``Learning low-dimensional
  representations of medical concepts,'' \emph{AMIA Summits on Translational
  Science Proceedings}, vol. 2016, p.~41, 2016.

\bibitem{Ma2018-gu}
F.~Ma, Q.~You, H.~Xiao, R.~Chitta, J.~Zhou, and J.~Gao, ``{KAME}:
  Knowledge-based attention model for diagnosis prediction in healthcare,'' in
  \emph{CIKM}.\hskip 1em plus 0.5em minus 0.4em\relax ACM, Oct. 2018, pp.
  743--752.

\bibitem{Peng2019TemporalSN}
X.~Peng, G.~Long, T.~Shen, S.~Wang, J.~Jiang, and M.~Blumenstein, ``Temporal
  self-attention network for medical concept embedding,'' in \emph{ICDM}.\hskip
  1em plus 0.5em minus 0.4em\relax IEEE, 2019, pp. 498--507.

\bibitem{peng2020self}
X.~Peng, G.~Long, T.~Shen, S.~Wang, and J.~Jiang, ``Self-attention enhanced
  patient journey understanding in healthcare system,'' \emph{arXiv preprint
  arXiv:2006.10516}, 2020.

\bibitem{Ma2018-ao}
F.~Ma, J.~Gao, Q.~Suo, Q.~You, J.~Zhou, and A.~Zhang, ``Risk prediction on
  electronic health records with prior medical knowledge,'' in
  \emph{SIGKDD}.\hskip 1em plus 0.5em minus 0.4em\relax ACM, Jul. 2018, pp.
  1910--1919.

\bibitem{ma2018health}
T.~Ma, C.~Xiao, and F.~Wang, ``Health-atm: A deep architecture for multifaceted
  patient health record representation and risk prediction,'' in
  \emph{SDM}.\hskip 1em plus 0.5em minus 0.4em\relax SIAM, 2018, pp. 261--269.

\bibitem{nguyen2016deepr}
P.~Nguyen, T.~Tran, N.~Wickramasinghe, and S.~Venkatesh, ``Deepr: A
  convolutional net for medical records,'' 2016.

\bibitem{shang2015neural}
L.~Shang, Z.~Lu, and H.~Li, ``Neural responding machine for short-text
  conversation,'' \emph{arXiv:1503.02364}, 2015.

\bibitem{sukhbaatar2015end}
S.~Sukhbaatar, J.~Weston, R.~Fergus \emph{et~al.}, ``End-to-end memory
  networks,'' in \emph{NeurIPS}, 2015, pp. 2440--2448.

\bibitem{rush2015neural}
A.~M. Rush, S.~Chopra, and J.~Weston, ``A neural attention model for
  abstractive sentence summarization,'' \emph{arXiv:1509.00685}, 2015.

\bibitem{lin2017structured}
Z.~Lin, M.~Feng, C.~N.~d. Santos, M.~Yu, B.~Xiang, B.~Zhou, and Y.~Bengio, ``A
  structured self-attentive sentence embedding,'' \emph{arXiv:1703.03130},
  2017.

\bibitem{liu2016learning}
Y.~Liu, C.~Sun, L.~Lin, and X.~Wang, ``Learning natural language inference
  using bidirectional lstm model and inner-attention,''
  \emph{arXiv:1605.09090}, 2016.

\bibitem{shen2018bi}
T.~Shen, T.~Zhou, G.~Long, J.~Jiang, and C.~Zhang, ``Bi-directional block
  self-attention for fast and memory-efficient sequence modeling,''
  \emph{arXiv:1804.00857}, 2018.

\bibitem{hu2017reinforced}
M.~Hu, Y.~Peng, Z.~Huang, X.~Qiu, F.~Wei, and M.~Zhou, ``Reinforced mnemonic
  reader for machine reading comprehension,'' \emph{arXiv:1705.02798}, 2017.

\bibitem{he2016deep}
K.~He, X.~Zhang, S.~Ren, and J.~Sun, ``Deep residual learning for image
  recognition,'' in \emph{CVPR}, 2016, pp. 770--778.

\bibitem{ba2016layer}
J.~L. Ba, J.~R. Kiros, and G.~E. Hinton, ``Layer normalization,'' \emph{arXiv
  preprint arXiv:1607.06450}, 2016.

\bibitem{xiao2018readmission}
C.~Xiao, T.~Ma, A.~B. Dieng, D.~M. Blei, and F.~Wang, ``Readmission prediction
  via deep contextual embedding of clinical concepts,'' \emph{PloS one},
  vol.~13, no.~4, p. e0195024, 2018.

\bibitem{MCE_Cai_2018}
X.~Cai, J.~Gao, K.~Y. Ngiam, B.~C. Ooi, Y.~Zhang, and X.~Yuan, ``Medical
  concept embedding with time-aware attention,'' in \emph{IJCAI}, 2018, pp.
  3984--3990.

\end{thebibliography}

\end{document}